\documentclass[11pt,a4paper]{article}
\usepackage[hyperref]{acl2021}
\usepackage{times}
\usepackage{latexsym}

\usepackage{graphicx}
\usepackage{amsmath}
\usepackage{amsthm}
\usepackage{booktabs}
\usepackage{algorithm}
\usepackage{algorithmic}
\usepackage{multirow}
\usepackage{arydshln}
\usepackage{xcolor}

\usepackage{CJKutf8}

\usepackage{microtype}
\usepackage{arydshln}

\aclfinalcopy 
\def\aclpaperid{2474} 

\title{On the Effectiveness of Adapter-based Tuning for\\ Pretrained Language Model Adaptation}

\author{Ruidan He\thanks{$^{*}$ Equally Contributed}$^{~1}$, Linlin Liu$^{*}$$^{12}$\thanks{$^\dag$ Linlin, Qingyu, Bosheng, Liying, and Jia-wei are under the Joint PhD Program between Alibaba and their corresponding universities.}~, Hai Ye$^{*}$$^{3}$, Qingyu Tan$^{13}$$^\dag$, Bosheng Ding$^{12}$$^\dag$, \\ \textbf{Liying Cheng$^{14}$$^\dag$, Jia-Wei Low$^{12}$$^\dag$, Lidong Bing$^{1}$, Luo Si$^{1}$} 
\\$^1$DAMO Academy, Alibaba Group~~~
$^2$Nanyang Technological University\\
$^3$National University of Singapore ~~~
$^4$Singapore University of Technology and Design\\
\texttt{\{ruidan.he,linlin.liu,qingyu.tan\}@alibaba-inc.com}\\
\texttt{\{bosheng.ding,liying.cheng,jiawei.low\}@alibaba-inc.com}\\
\texttt{yeh@comp.nus.edu.sg}~~~\texttt{\{l.bing,luo.si\}@alibaba-inc.com}}

\date{}

\begin{document}
\maketitle
\begin{abstract}

Adapter-based tuning has recently arisen as an alternative to fine-tuning. It works by adding light-weight adapter modules to a pretrained language model (PrLM) and only updating the parameters of adapter modules when learning on a downstream task. As such, it adds only a few trainable parameters per new task, allowing a high degree of parameter sharing. Prior studies have shown that adapter-based tuning often achieves comparable results to fine-tuning. However, existing work only focuses on the parameter-efficient aspect of adapter-based tuning while lacking further investigation on its effectiveness. In this paper, we study the latter. 
We first show that adapter-based tuning better mitigates forgetting issues than fine-tuning since 
it yields representations with less deviation from those generated by the initial PrLM.
We then empirically compare the two tuning methods on several downstream NLP tasks and settings. We demonstrate that 1) adapter-based tuning outperforms fine-tuning on \emph{low-resource} and \emph{cross-lingual} tasks; 2) it is more robust to overfitting and \emph{less sensitive} to changes in learning rates.

\end{abstract}

\section{Introduction}\label{sec:intro}

Large scale pretrained language models~(PrLMs) \citep{devlin2019bert,liu2019roberta,conneau2020unsupervised,brown2020language} have achieved state-of-the-art results on most natural language processing (NLP) tasks, where fine-tuning has become a dominant approach to utilize PrLMs. A standard fine-tuning process copies weights from a PrLM and tunes them on a downstream task, which requires a new set of weights for each task.

Adapter-based tuning \citep{houlsby2019a,bapna2019simple} has been proposed as a more parameter-efficient alternative. 
For NLP, adapters are usually light-weight modules inserted between transformer layers \citep{vaswani2017}. During model tuning on a downstream task, only the parameters of adapters are updated while the weights of the original PrLM are frozen. 
Hence, adapter-based tuning adds only a small amount of parameters for each task, allowing a high degree of parameter-sharing. Though using much less trainable parameters, adapter-based tuning has demonstrated comparable performance with full PrLM fine-tuning \citep{houlsby2019a,bapna2019simple,stickland2019a}.

Existing work mostly focuses on the parameter-efficient aspect of adapters and attempt to derive useful applications from that, which is still the case in most recent works: \citet{ruckle2020adapterdrop} explore methods to further improve the parameter and computation efficiency of adapters; \citet{pfeiffer2020adapterfusion} combine knowledge from multiple adapters to improve the performance on downstream tasks; \citet{artetxe2020cross} and \citet{pfeiffer2020madx} leverage the modular architecture of adapters for parameter-efficient transfer to new languages or tasks, and \citet{wang2020kadapter} utilize the same property for knowledge injection.

Besides parameter-efficiency, the unique characteristic of adapter-based tuning, with alternating frozen and learnable layers, might be directly useful for improving model performances. However, this has not yet been discussed in the prior work.  In this paper, we first empirically demonstrate that adapter-based tuning better regularizes training than fine-tuning by mitigating the issue of forgetting. We show that it yields representations with less deviation from those generated by the original PrLM. 
Next, to see what this property of adapters will help when adapting PrLMs, we compare the performance of fine-tuning and adapter-based tuning on a wide range of datasets and NLP tasks. Extensive experiments and analysis are conducted in different settings, including low-resource and high-resource, monolingual and cross-lingual. 

Our main findings can be summarized as follows:
\begin{itemize}
    \item For monolingual adaptation, adapter-based tuning yields better results in \emph{low-resource settings}, especially when the task is more domain-specific. With increasing training samples, the performance gain over fine-tuning is less significant~($\S$\ref{sec:text_clf}).
    \item Adapter-based tuning tends to outperform fine-tuning on \emph{zero-shot cross-lingual tasks} under different amounts of training data~($\S$\ref{sec:cross-lingual}).
    \item Adapter-based tuning demonstrates higher stability and better generalization ability. It is less sensitive to learning rates compared to fine-tuning~($\S$\ref{sec:analysis}).
\end{itemize}

\begin{figure}
\centering
\includegraphics[width=0.9\columnwidth]{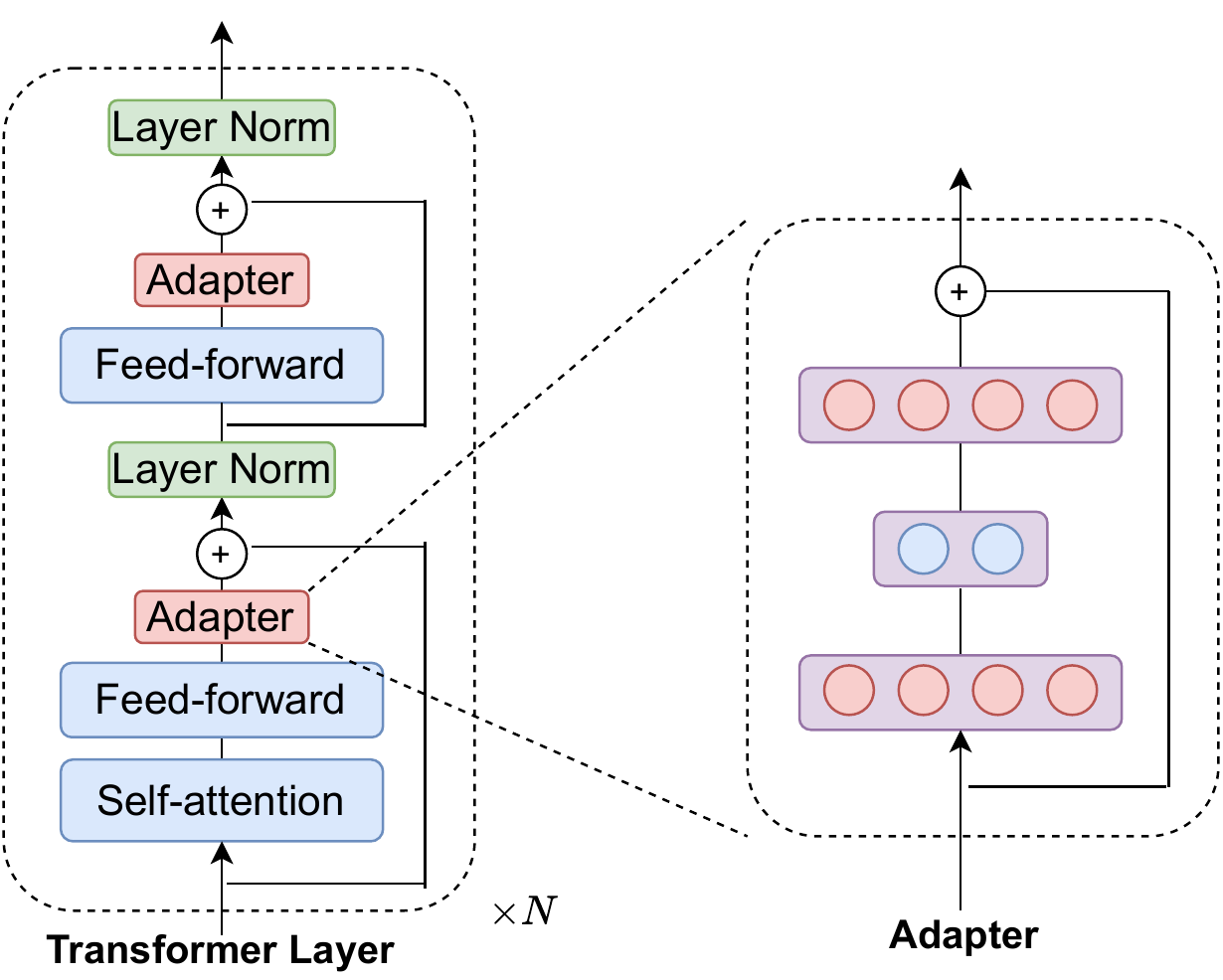}
\caption{The structure of the adapter adopted from~\citet{houlsby2019a}. $N$ is the number of transformer layers.}
\label{fig:adapter}
\end{figure}

\section{Adapter Better Regularizes Tuning}\label{sec:model}
\subsection{Adapter-based Tuning}\label{sec:background}

\begin{figure}
\centering
\includegraphics[width=0.85\columnwidth]{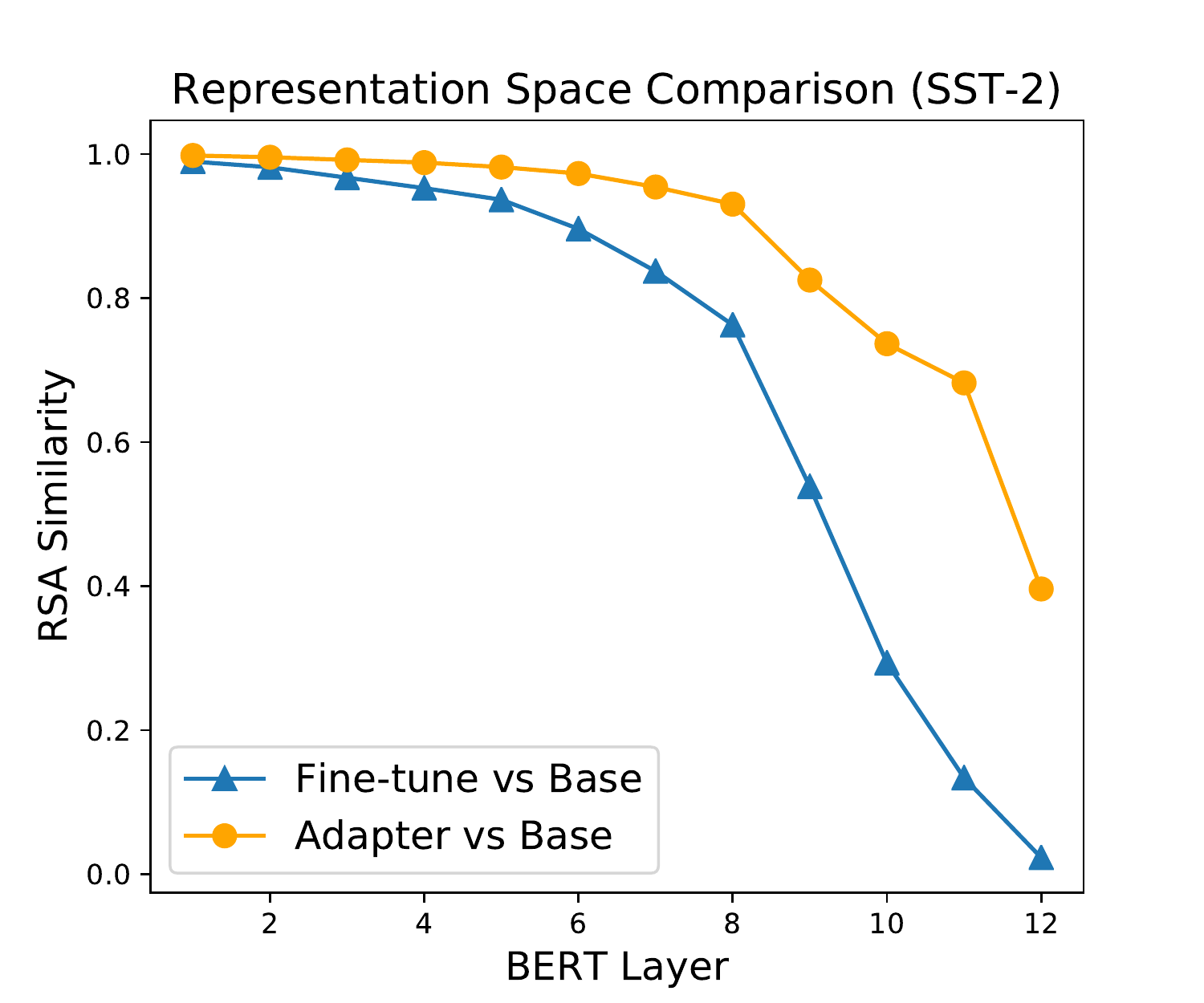}
\caption{Comparison of the representations obtained at each layer before (\emph{Base}) and after adapter-based tuning or fine-tuning on BERT-base using Representational Similarity Analysis (RSA). 5000 tokens are randomly sampled from the dev set for computing RSA.  A higher score indicates that the representation spaces before and after tuning are more similar.}
\label{fig:rsa_sst2}
\end{figure}

When adapting a pretrained language model~(PrLM), adapter-based tuning inserts light-weight neural networks (adapters) between the transformer layers of the PrLM, and only updates the parameters of the adapters on a downstream task, but keeps the ones of the PrLM frozen. 
Unlike fine-tuning which introduces an entire new model for every task, one great advantage of adapter-based tuning is generating a compact model with only a few trainable parameters added per task. 

\citet{houlsby2019a} have extensively studied the choices of adapter architectures and where they should be inserted into PrLMs. They find that a stack of down- and up-scale neural networks works well which only introduces a small amount of extra parameters to the network. This design inspires most of the following work~\cite{pfeiffer2020adapterfusion,pfeiffer2020madx,bapna2019simple}. As shown in Figure~\ref{fig:adapter}, the adapter maps an input hidden vector $\mathrm{h}$ from dimension $d$ to dimension $m$ where $m<d$, and then re-maps it to dimension $d$. We refer $m$ as the \textbf{hidden size} of the adapter. A skip-connection is employed inside the adapter network such that if the parameters of the projection layers are near zeros, the adapter module approximates an identity function. Formally, given the input hidden vector $\mathrm{h}$, the output vector $\mathrm{h}'$ is calculated as:
\begin{equation}
    \mathrm{h}' =  f_2(\tanh{f_1(\mathrm{h})}) +  \mathrm{h}
\end{equation}
in which $f_1(\cdot)$ and $f_2(\cdot)$ are the down- and up-projection layers. At each transformer layer, two adapters are inserted right after the self-attention and the feed-forward layers respectively. During adapter tuning, only the parameters of the adapters, the normalization layers, and the final classification layer are updated. We use the above described adapter configuration in all of our experiments, since it is adopted in most prior work with few modifications.

\subsection{Representation Similarity}

Fine-tuning large-scale PrLMs on downstream tasks can suffer from overfitting and bad generalization issues~\cite{DBLP:journals/corr/abs-2002-06305, DBLP:journals/corr/abs-1811-01088}. Recently, \citet{DBLP:conf/iclr/LeeCK20} propose Mixout to regularize the fine-tuning of PrLMs. They show that Mixout avoids catastrophic forgetting and stabilizes the fine-tuning process by encouraging the weights of the updated model to stay close to the initial weights. Since adapter-based tuning does not update the weights of PrLMs at all, we suspect that it has a similar effect of alleviating the issue of catastrophic forgetting. Since the weights of the PrLM are the same before and after adapter-based tuning, to verify this, we use Representational Similarity Analysis (RSA)~\cite{laakso2000}
to assess the similarity of tuned representations to those without tuning at each transformer layer. 

RSA has been widely used to analyze the similarity between two neural network outputs~\cite{Abnar2019, chrupala2019, merchant2020}, which works by 
creating two comparable sets of representations by inputting a same set of $n$ samples to the two models. 
For each set of representations, a $n \times n$ pairwise similarity\footnote{Cosine similarity is used} matrix is calculated. The final RSA similarity score between the two representation space is computed as the Pearson correlation between the flattened upper triangulars of the two similarity matrices. We use a subset of GLUE tasks~\cite{wang2018} for our analysis. Given a task, we first perform adapter-based tuning and fine-tuning to adapt a BERT-base model ($M_{org}$) to the target task, which yields models $M_{adapt}$ and $M_{ft}$ respectively (See Appendix~\ref{append: experimental setup} for training details). Then we pass sentences (or sentence-pairs depend on the task) from the development set to $M_{org}$, $M_{adapt}$, and $M_{ft}$ respectively. 
We extract representations at each layer from the three models and select the corresponding representations of 5k randomly sampled tokens\footnote{We skip [PAD], [CLS], [SEP] for token selection.} ($n=5000$) for evaluation. Note that the same set of tokens is used for all models. Finally, we compare the representations obtained from $M_{adapt}$ or $M_{ft}$ to those from $M_{org}$ using RSA.

Figure~\ref{fig:rsa_sst2} plots the results on STS-2, results of other tasks demonstrate a similar trend and can be found in Appendix~\ref{append: additional_results}. For both fine-tuning and adapter-based tuning, we observe that the representation change generally arises in the top layers of the network, which is consistent with previous findings that higher layers are more task relevant~\cite{DBLP:conf/acl/RuderH18}. It can be clearly observed that compared to fine-tuning, adapter-based tuning yields representations with less deviation from those of BERT-base at each layer, which verifies our claim that adapter-based tuning can better regularize the tuning process by mitigating the forgetting problem. Apparently, this property of adapter tuning comes from that it freezes all the parameters of PrLMs. And because of the skip-connection in the adapter, the hidden representation out of the adapter can mimic the input representation, in this way, some of the original knowledge of PrLMs~(before injecting adapters) can be preserved. 

Since we find that adapter-based tuning better regularizes the learning process, the next question is how this property will help to improve the performance when adapting PrLMs to downstream tasks. We conduct extensive experiments to investigate this. The remainder of this paper is organized as follows. We compare fine-tuning and adapter-based tuning on monolingual text-level adaptation tasks in $\S$\ref{sec:text_clf}, followed by cross-lingual adaptation in $\S$\ref{sec:cross-lingual}. Further analysis about the training stability and generalization capabilities is shown in $\S$\ref{sec:analysis}.

\section{Monolingual Adaptation}\label{sec:text_clf}

In this section, we first experiment with eight datasets as used in \citet{gururangan2020} including both high- and low-resource tasks ($\S$\ref{sec:tapt}). We refer this set of tasks as \emph{Task Adaptation Evaluation}~(\textbf{TAE}). We observe that adapter-based tuning consistently outperforms fine-tuning on low-resource tasks, while they perform similarly on high-resource tasks. We further confirm the effectiveness of adapters in low-resource settings on the GLUE benchmark~\cite{wang2018}~($\S$\ref{sec:glue}).

\subsection{TAE}\label{sec:tapt}

\begin{table*}[t]
\centering
\resizebox{\textwidth}{!}{
\begin{tabular}{lcccc@{\hspace{1.5cm}}cccc}
\toprule
&\multicolumn{4}{c}{\bf{low-resource}} &\multicolumn{4}{c}{\bf{high-resource}}\\
\multirow{2}*{\bf{Model}} &\bf{CHEMPROT} &\bf{ACL-ARC} &\bf{SCIERC} &\bf{HYP.} &\bf{RCT} &\bf{AGNEWS} &\bf{HELPFUL.} &\bf{IMDB}\\
&(4169) &(1688) &(3219) &(515) &(180k) &(115k) &(115k)  &(20k)\\
\midrule
RoBa.-ft$^{\dag}$&81.9$_{1.0}$ &63.0$_{5.8}$ &77.3$_{1.9}$ &86.6$_{0.9}$ &\bf{87.2}$_{0.1}$ &\bf{93.9}$_{0.2}$ &65.1$_{3.4}$ &95.0$_{0.2}$ \\
RoBa.-ft$^{*}$&81.7$_{0.8}$ & 65.0$_{3.6}$& 78.5$_{1.8}$& 88.9$_{3.3}$ & 87.0$_{0.1}$& 93.7$_{0.2}$& \bf{69.1}$_{0.6}$ & 95.2$_{0.1}$\\
RoBa.-adapter$_{256}$ & \bf{82.9}$_{0.6}$  &\bf{67.5}$_{4.3}$& \bf{80.8}$_{0.7}$ & \bf{90.4}$_{4.2}$ & 87.1$_{0.1}$ &93.8$_{0.1}$& 69.0$_{0.4}$ & \bf{95.7}$_{0.1}$\\
\midrule

RoBa.-ft+TAPT$^{\dag}$&82.6$_{0.4}$ &67.4$_{1.8}$ &79.3$_{1.5}$ &\bf{90.4}$_{5.2}$  &\bf{87.7}$_{0.4}$ &\bf{94.5}$_{0.1}$ &68.5$_{1.9}$ &95.5$_{0.1}$\\
RoBa.-ft+TAPT$^{*}$&82.5$_{0.3}$ & 66.5$_{5.1}$& 79.7$_{0.8}$ & 91.3$_{0.8}$  & 87.4$_{0.1}$ & 94.0$_{0.2}$& \bf{70.3}$_{1.1}$ & 95.4$_{0.1}$\\
RoBa.-adapter$_{256}$+TAPT &\bf{83.5}$_{0.5}$ & \bf{70.0}$_{2.1}$& \bf{81.1}$_{0.2}$&90.0$_{3.5}$ & 87.2$_{0.1}$ & 94.0$_{0.1}$& 68.8$_{0.8}$ & \bf{95.8}$_{0.0}$\\
\bottomrule
\end{tabular}}
\caption{Average results across five random seeds with standard deviations as subscripts on TAE. micro-F1 is reported for CHEMPROOT and RCT, and macro-F1 is reported for the other tasks.  Results with ``$^{\dag}$'' are taken from \citet{gururangan2020}. Results with ``*'' are reproduced by us. Numbers in () indicate the training size.} \label{table: tapt_main_comparison}
\end{table*}

TAE consists of four domains~(biomedical, computer science, news text, and AMAZON reviews) and eight classification tasks~(two in each domain), whose domain diversity makes it suitable to assess the adaptation effectiveness of different approaches. Detailed data statistics are displayed in Appendix~\ref{append:data}. We consider tasks with fewer than 5k training examples as low-resource tasks and the others as high-resource tasks.

\paragraph{Experimental Setup} 
We perform supervised fine-tuning on RoBERTa-base as our baseline (\textbf{RoBa.-ft}). For adapter-based tuning, we set the hidden size $m$ of adapters to 256~(\textbf{RoBa.-adapter$_{256}$}). 
We also present the results of adding \emph{task-adaptive pretraining}~(\textbf{+TAPT})~\cite{gururangan2020}. In this setting, before fine-tuning or adapter-based tuning, the model was trained with a masked language modeling~(MLM) objective on the training texts (without labels) of the task. Note that in \emph{RoBa.-adapter$_{256}$+TAPT}, we also use adapter-based tuning for TAPT where only the weights of adapters are updated at the TAPT stage. This is to evaluate whether adapter-based tuning can work with unsupervised learning objectives. We follow the experimental settings in \citet{gururangan2020} for TAPT. For fine-tuning and adapter-based tuning, we train models for 20 epochs to make sure they are sufficiently trained and save the checkpoint after each training epoch. We select the checkpoint that achieves the best score on the validation set for evaluation on the test set. The batch size is set to 16 for both methods. The learning rate is set to 2e-5 for fine-tuning, and 1e-4 for adapter-based tuning. 
See Appendix~\ref{append: experimental setup} for the hyperparameter selection process and more training details.

\begin{figure}
\centering
\includegraphics[width=\columnwidth]{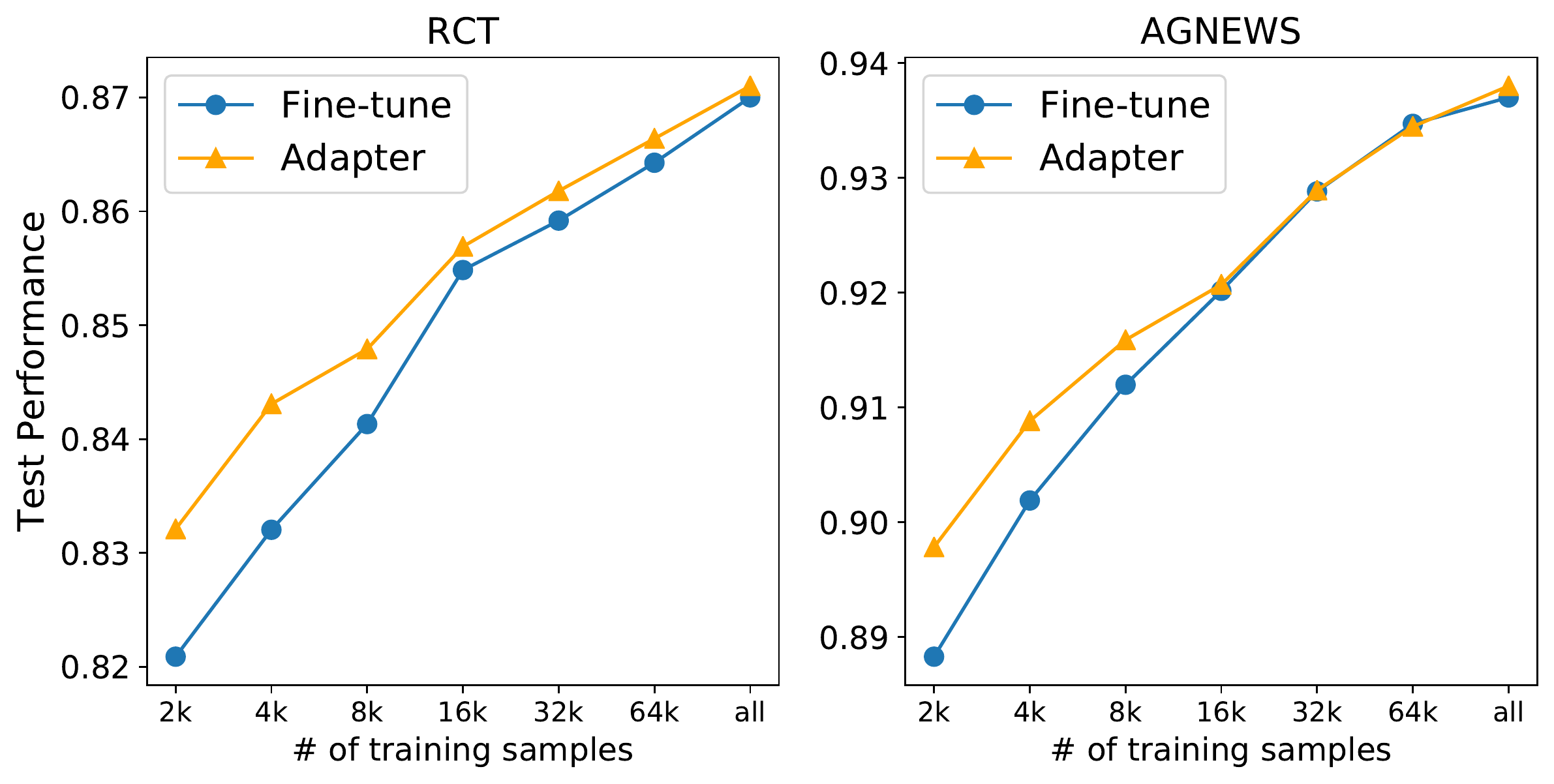}
\caption{Test performance w.r.t the number of training examples. Reported results are averages across five runs with different random seeds.}\label{fig:tapt_vary_size}
\end{figure}

\paragraph{Results} 
Table~\ref{table: tapt_main_comparison} presents the comparison results. We report the average result over 5 runs with different random seeds. On four low-resource tasks, adapter-based tuning consistently outperforms fine-tuning and improves the average result by 1.9\%. Adapter-based tuning alone without TAPT even outperforms fine-tuning with TAPT. Besides, adding TAPT before adapter-based tuning further improves the performance on 3 out of 4 low-resource tasks, which suggests that adapter-based tuning works with both supervised and unsupervised objectives. 
Another finding is that when trained on high-resource tasks, both methods achieve similar results. To verify the effects of training size, on high-resource tasks, we plot the performances  with varying numbers of training examples in Figure~\ref{fig:tapt_vary_size}. The trend is consistent with our existing observations -- adapter-based tuning achieves better results when the training set is small while fine-tuning will gradually catch up with an increasing number of training examples.

\subsection{GLUE Low-resource Adaptation}\label{sec:glue}
\begin{table*}[t]
\centering
\resizebox{\textwidth}{!}{
\begin{tabular}{lcccccccccc}
\toprule
\textbf{Model} &\bf{CoLA} &\bf{MNLI$_m$} &\bf{MNLI$_{mm}$} &\bf{MRPC} &\bf{QNLI} &\bf{QQP} &\bf{RTE} &\bf{SST-2} &\bf{STS-B} &\bf{Avg.}\\
\midrule
\multicolumn{11}{l}{\textbf{1k}}\\
BERT-ft&41.4$_{4.0}$& 57.4$_{3.2}$& 60.3$_{3.2}$& 83.6$_{1.2}$& 80.5$_{0.3}$& 69.8$_{0.7}$& 62.5$_{1.1}$& 87.8$_{0.4}$& 85.5$_{0.9}$& 69.9$_{1.7}$\\
BERT-adapter$_{64}$ &42.9$_{2.6}$& 61.6$_{0.9}$& 64.1$_{0.8}$& 84.8$_{0.7}$& 80.5$_{0.9}$& 70.3$_{2.0}$& 62.5$_{1.3}$& 88.0$_{0.7}$& 86.1$_{0.3}$& 71.2$_{1.1}$\\
BERT-adapter$_{64-256}$ &\bf{43.6}$_{2.9}$& \bf{61.6}$_{0.9}$& \bf{64.1}$_{0.8}$& \bf{84.8}$_{0.7}$& \bf{81.0}$_{0.2}$& \bf{76.8}$_{0.7}$& \bf{65.3}$_{2.0}$& \bf{88.0}$_{0.7}$& \bf{86.3}$_{0.2}$& \bf{72.4}$_{1.0}$\\
\\

RoBa.-ft&45.4$_{2.8}$& 71.2$_{0.9}$& 72.9$_{0.9}$& 88.4$_{0.7}$& \bf{84.0}$_{0.7}$& 75.0$_{1.1}$& 67.0$_{2.7}$& 89.0$_{0.8}$& 88.5$_{0.4}$& 75.7$_{1.2}$\\
RoBa.-adapter$_{64}$ &47.7$_{2.5}$& 71.0$_{0.8}$& 71.9$_{0.8}$& 88.9$_{0.9}$& 83.2$_{0.5}$& 74.7$_{0.3}$& 67.7$_{2.2}$& 90.0$_{1.4}$& 88.4$_{0.2}$& 76.0$_{1.1}$\\
RoBa.-adapter$_{64-256}$ &\bf{47.7}$_{2.5}$& \bf{71.8}$_{0.8}$& \bf{73.0}$_{1.1}$& \bf{89.2}$_{0.7}$& 83.5$_{0.4}$& \bf{75.1}$_{0.1}$& \bf{68.7}$_{0.8}$& \bf{90.5}$_{0.2}$& \bf{88.6}$_{0.2}$& \bf{76.4}$_{0.8}$\\

\midrule
\multicolumn{11}{l}{\textbf{5k}}\\
BERT-ft&\bf{54.4}$_{2.4}$& 69.6$_{0.8}$& 71.2$_{1.1}$& -& 85.0$_{0.7}$& 74.7$_{1.8}$& -& 88.6$_{1.0}$& 88.7$_{0.7}$& 76.0$_{1.2}$\\
BERT-adapter$_{64}$ &54.1$_{1.5}$& 71.3$_{0.5}$& 73.0$_{0.4}$& -& 85.3$_{0.3}$& 74.2$_{1.3}$& -& 89.1$_{0.2}$& 88.9$_{0.1}$& 76.6$_{0.6}$\\
BERT-adapter$_{64-256}$ &54.1$_{1.5}$& \bf{71.3}$_{0.5}$& \bf{73.2}$_{0.4}$& -& \bf{85.3}$_{0.3}$& \bf{74.9}$_{0.4}$& -& \bf{89.1}$_{0.2}$& \bf{88.9}$_{0.1}$& \bf{76.7}$_{0.5}$\\
\\

RoBa.-ft&55.7$_{1.7}$& 79.5$_{0.4}$& 80.3$_{0.4}$& -& \bf{87.1}$_{0.5}$& 78.1$_{1.3}$& -& 91.4$_{0.5}$& \bf{90.6}$_{0.1}$& 80.4$_{0.7}$\\
RoBa.-adapter$_{64}$ &56.8$_{1.2}$& 80.2$_{0.3}$& 80.6$_{0.2}$& -& 86.5$_{0.7}$& 78.2$_{1.0}$& -& 92.2$_{0.5}$& 90.4$_{0.2}$& 80.7$_{0.6}$\\
RoBa.-adapter$_{64-256}$ &\bf{57.4}$_{1.6}$& \bf{80.2}$_{0.3}$& \bf{80.5}$_{0.2}$& -& 86.9$_{0.6}$& \bf{78.3}$_{0.9}$& -& \bf{92.2}$_{0.5}$& 90.4$_{0.2}$& \bf{80.8}$_{0.6}$\\
\bottomrule
\end{tabular}}
\caption{Results on GLUE 1k and 5k low resource settings as described in $\S$\ref{sec:glue}. Results of MRPC and RTE in 5k setting are omitted as their training data is less than 5k. CoLA is evaluated using Matthew's Correlation. MRPC and QQP are evaluated using F1 score. STS-B is evaluated using Spearman's correlation.  The other tasks are evaluated using accuracy. We report averages across five random seeds, with standard deviations as subscripts.} \label{table: glue_main_comparison}
\end{table*}

To further validate that adapters tend to generalize better than fine-tuning under low-resource settings, we follow \citet{DBLP:journals/corr/abs-2006-05987} to study low-resource adaptation using eight datasets from the GLUE benchmark~\cite{wang2018} which covers four types of tasks: natural language inference (MNLI, QNLI, RTE), paraphrase detection (MRPC, QQP), sentiment classification (SST-2) and linguistic acceptability (CoLA). Appendix~\ref{append:data} provides detailed data statistics and descriptions. 

\paragraph{Experimental Setup} 
For each dataset, we simulate two low-resource settings by randomly sampling 1k and 5k instances from the original training data as the new training sets.  In each setting, we draw another 1k samples from the remaining training set as the validation set and instead use the original validation set as the test set, since the original GLUE test sets are not publicly available~\footnote{Users are limited to a maximum of two submissions per day to obtain test results, which is inconvenient for a large number of runs}. 

We perform fine-tuning on BERT-base (\textbf{BERT-ft}) and RoBERTa-base (\textbf{RoBa.-ft}) respectively as our baselines. 
We set the learning rate to 2e-5 and the batch size to 16 for BERT and RoBERTa fine-tuning experiments~(See Appendix~\ref{append: experimental setup} for details). For adapters, we only tune its hidden sizes in \{64, 128, 256\},  setting the learning rate to 1e-4 and batch size to 16 as the same used in $\S$\ref{sec:tapt}. 

\paragraph{Results} 
Table~\ref{table: glue_main_comparison} presents the comparison results. For adapter-based tuning, we report two results on each task. One is obtained with the optimal hidden size which varies per dataset, and the other is obtained with the size of 64. We observe that adapter-based tuning outperforms fine-tuning most of the time under both 1k and 5k settings. In particular, the performance gain is more significant in 1k setting, where on average across all tasks, adapter-based tuning outperforms fine-tuning by 2.5\% and 0.7\% on BERT and RoBERTa respectively.

\subsection{Discussions}

One consistent observation from $\S$~\ref{sec:tapt} and $\S$~\ref{sec:glue} is that adapters tend to outperform fine-tuning on text-level classification tasks when the training set is small, but with more training samples, the benefit of adapters is less significant. In low-resource setting, fine-tuning has more severe overfitting problem, since it has much more tunable parameters compared to adapter-tuning, so adapter-tuning works better than fine-tuning. However, in high-resource setting, overfitting is not a big issue and model capacity counts more. Obviously, the model capacity under fine-tuning is larger than that under adapter-tuning since fine-tuning can update much more model parameters. 



When comparing the improvements of adapter tuning over fine-tuning on tasks from TAE~($\S$~\ref{sec:tapt}) and GLUE~($\S$~\ref{sec:glue}), we find that the improvement is more significant on low-resource tasks from TAE -- on RoBERTa-base, the average improvement brought by adapters is 1.9\% across four low-resource tasks from TAE, while the average improvement on GLUE is 0.7\% and 0.4\% in 1k and 5k settings respectively. As indicated in \citet{gururangan2020}, the TAE dataset is more domain-specific and has less overlap with the corpus used for RoBERTa-base pretraining, one intuitive explanation for this observation is that fine-tuning has more severe forgetting and overfitting issues in domain adaptation where the target domain is dissimilar to the source domain in pretraining, thus adapter-based tuning is more preferable in this scenario. 

\begin{table*}[t]
\centering
\scalebox{0.75}{
\begin{tabular}{lccc@{\hspace{1.2cm}}ccc@{\hspace{1.2cm}}ccc}
\toprule
&&\bf{POS}&&&\bf{NER}&&&\bf{XNLI}&\\
\textbf{Model} &\textbf{All} &\textbf{Target} &\textbf{Distant} &\textbf{All} &\textbf{Target} &\textbf{Distant} &\textbf{All} &\textbf{Target} &\textbf{Distant}\\
\midrule
XLMR-ft~\cite{hu2020b} &73.80 &73.14 &64.34 &65.40 &64.87 &58.21 &79.24 &78.56 &76.73\\
XLMR-ft (reproduced) &74.29 &73.61 &64.90 &63.85 &63.32 &56.85 &79.28 &78.64 &77.03\\
XLMR-adapter$_{256}$ &\bf{75.82} &\bf{75.20} &\bf{68.05} &\bf{66.40} &\bf{65.95} &\bf{59.01} &\bf{80.08} &\bf{79.43} &\bf{77.60}\\
\bottomrule
\end{tabular}}
\caption{Zero-shot cross-lingual results.  Accuracy is reported for POS tagging and XNLI. F1 is reported for NER. \textbf{All} is the average test result of all languages. \textbf{Target} is the average test result of all target languages except English. \textbf{Distant} is the average test result of the languages not in the Indo-European family.} \label{table: cross_lingual_main}
\end{table*}
\begin{table*}[!]
\centering
\scalebox{0.75}{
\begin{tabular}{lccc@{\hspace{1.2cm}}ccc@{\hspace{1.2cm}}ccc}
\toprule

&&\bf{5\%}&&&\bf{10\%}&&&\bf{20\%}&\\
\textbf{Model} & \bf{All} & \bf{Target} & \bf{Distant} & \bf{All} & \bf{Target} & \bf{Distant} & \bf{All} & \bf{Target} & \bf{Distant}\\
\midrule
XLMR-ft & 75.76 & 75.09 & 73.12 & 76.73 & 76.07 & 74.21 & 78.28 & 77.64 & 75.84 \\

XLMR-adapter$_{64}$ & \textbf{76.09} & \textbf{75.47} & \textbf{73.78} & \textbf{77.52} & \textbf{76.94} & \textbf{75.10} & \textbf{78.68} & \textbf{78.07} & \textbf{76.39} \\
\bottomrule
\end{tabular}}
\caption{Accuracy on XNLI with different amount of training data. We only compare XLMR-ft to XLMR-adapter$_{64}$ in this set of experiments as XLMR-adapter$_{64}$ is more light-weight. }
\label{table:xnli_low_resource}
\end{table*}
\section{Cross-lingual Adaptation}\label{sec:cross-lingual}

In this section, we further compare fine-tuning and adapter-based tuning in the zero-shot cross-lingual transfer setting. All experiments in this section are based on XLM-R-large~\citep{conneau2020unsupervised}, a recent SOTA multilingual PrLM covering 100 languages. 
We conduct evaluations on a set of multilingual tasks from XTREME~\cite{hu2020b}, including Universal Dependencies v2.5 tree banks (UD-POS)~\cite{nivre2018ud}, Wikiann NER~\cite{DBLP:conf/acl/PanZMNKJ17}, and cross-lingual natural language inference~(XNLI)~\cite{conneau2020xnli}. UD-POS contains 34 languages, Wikiann NER contains 40 languages, and XNLI contains 15 languages. We refer the reader to \citet{hu2020b} for additional details about the datasets. 

\paragraph{Experimental Setup}
On each task, we perform hyperparameter tuning on the English development set. For both fine-tuning and adapter-based tuning, we use batch size 32, and tune the learning rates in \{1e-5, 2e-5, 3e-5, 4e-5, 5e-5\}. For adapter-based tuning, we further tune the hidden sizes in \{64, 128, 256\} and find size 256 often performs the best. We train and select models with the English training and development sets and then evaluate the tuned models on test sets of all languages. See Appendix~\ref{append: experimental setup} for hyperparameter and training details. 

\paragraph{Results}
Table~\ref{table: cross_lingual_main} summarizes the results. 
To better compare cross-lingual transfer to different groups of languages, we present the average results of all languages~(\textbf{All}), the target languages except English~(\textbf{Target}), and the Non-Indo-European languages~(\textbf{Distant}). It can be observed that adapter-based tuning significantly outperforms fine-tuning on all three settings for each task. Specifically, adapter-based tuning outperforms the reported fine-tuning results~\cite{hu2020b} on \emph{Target} and \emph{Distant} by 2.06\% and 3.71\% on UD-POS, 1.08\% and 0.8\% on Wikiann NER, and 0.87\% and 0.87\% on XNLI. See Appendix~\ref{append: additional_results} for detailed results on each language.

Note that UD-POS, Wikiann NER, and XNLI are all high-resource tasks, with 20k, 20k, and 400k training samples respectively.
Unlike monolingual tasks, adapters achieve consistent performance gains even under high-resource settings on cross-lingual tasks. We suspect that the ability to mitigate forgetting is more useful in cross-lingual scenarios since the model knowledge of the target languages only comes from pretraining. Adapter-based tuning can better maintain the knowledge. 
We further investigate the effectiveness of adapter-based tuning on XNLI with smaller training sets. Table~\ref{table:xnli_low_resource} summarizes the results when trained on 5\%, 10\%, and 20\% of the original training sets. In all settings, adapters still demonstrate consistent improvements over fine-tuning.

\begin{figure*}[t]
\setlength{\abovecaptionskip}{-0.1cm}
\setlength{\belowcaptionskip}{-0.5cm}
\begin{center}
\includegraphics[width=\textwidth]{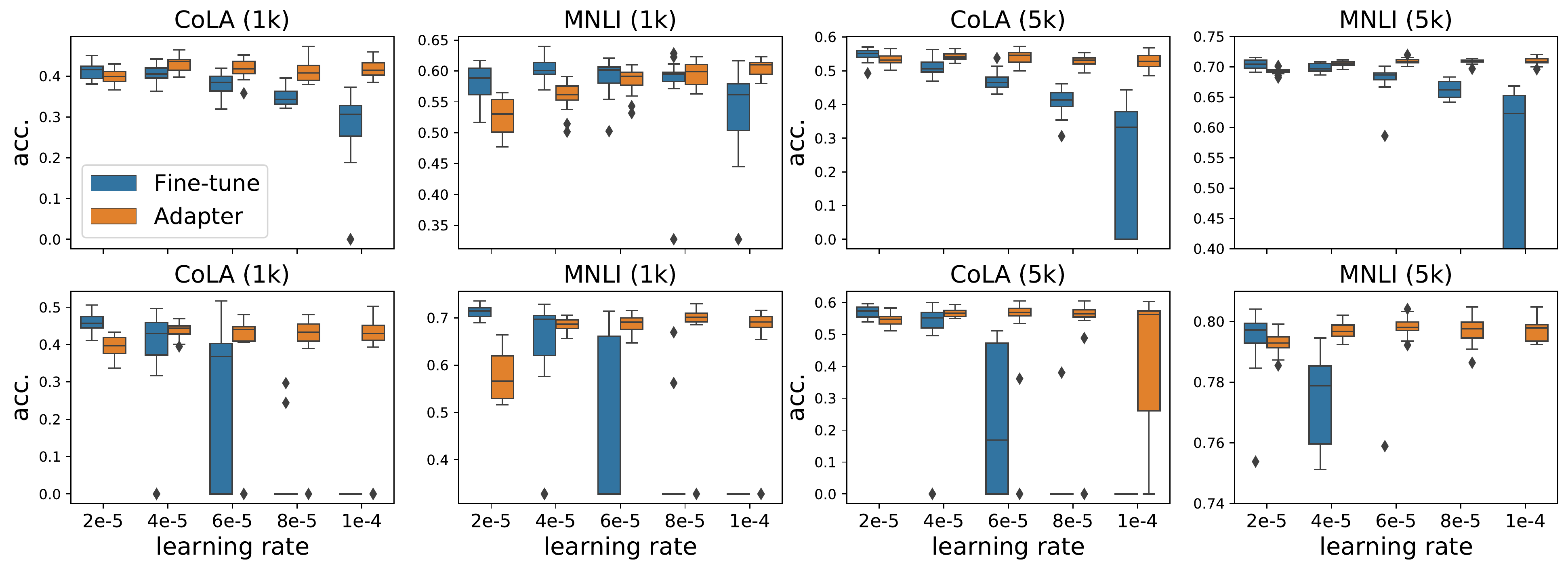}
\end{center}
\caption{Box plots of test performance distribution over 20 runs across different learning rates. The upper/bottom results are based on Bert-base/RoBERETa-base. Note that the fine-tuning results with learning rates larger than 4e-5 on RoBERTa. MNLI 5k are all zeros, which are outside of the range and not shown in the subplot. }
\label{fig:gamma}
\end{figure*}

\begin{table}[t]
\centering
\scalebox{0.75}{
\begin{tabular}{lcccc}
\toprule
\textbf{Model} &\bf{TAE$_{low}$} &\bf{GLUE$_{1k}$} &\bf{XNLI$_{full}$} &\bf{XNLI$_{5\%}$}\\
\midrule
finetune&78.52 &69.86& 78.64 & 75.09\\
Adapter$_{64}$&77.20 &\bf{71.20} & 79.01 & 75.47 \\
Adapter$_{128}$&79.29 &71.09 & 79.24 & \bf{75.83} \\
Adapter$_{256}$&\bf{80.41} &71.06 & \bf{79.43} & 75.45 \\
\bottomrule
\end{tabular}}
\caption{Average test results with different adapter hidden sizes. Results of GLUE$_{1k}$ are based on BERT-base. TAE$_{low}$ denotes low resource tasks from TAE.} \label{table:vary_hidden_size}
\end{table}

\section{Analysis}\label{sec:analysis}

\paragraph{Adapter Hidden Size}
The hidden size $m$\footnote{The fraction of adapter parameters w.r.t. BERT-base (110M parameters) is 2\%, 4\%, and 6\% when $m$ is set to 64, 128, and 256. The fraction w.r.t. XLMR-large (550M parameters) is 1\%, 2\%, and 3\%, respectively.} is the only adapter-specific hyperparameter. As indicated in \citet{houlsby2019a}, the hidden size provides a simple means to trade off performance with parameter efficiency. Table~\ref{table:vary_hidden_size} shows the performance with different hidden sizes, from which we find that increasing the hidden size may not always lead to performance gains. For monolingual low-resource adaptation, TAE tasks prefer a larger hidden size, while the results on GLUE are similar across different hidden sizes. We suspect that this is due to that TAE datasets are more dissimilar to the pretraining corpus, which requires relatively more trainable parameters to learn the domain-specific knowledge. On XNLI, a larger hidden size helps improve the performance when the full data is used. However, when only 5\% training data is used, increasing the hidden size does not yield consistent improvements. The results indicate that the optimal hidden size depends on both the domain and the training size of the task.

\paragraph{Learning Rate Robustness}
We compare the two tuning methods in terms of their stability w.r.t the learning rate. Figure~\ref{fig:gamma} shows the performance distributions on CoLA and MNLI under 1k and 5k settings. The learning rates are varied in \{2e-5, 4e-5, 6e-5, 8e-5, 1e-4\}. Each box in the plot is drawn from the results of 20 runs with different random seeds. We observe that fine-tuning yields larger variances when increasing the learning rates. It often collapses with learning rates larger than 4e-5 when RoBERTa-base is used. Adapter-based tuning is more stable across a wider range of learning rates.

\begin{figure}[t]
\setlength{\abovecaptionskip}{-0.1cm}
\setlength{\belowcaptionskip}{0.1cm}
\begin{center}
\includegraphics[width=\columnwidth]{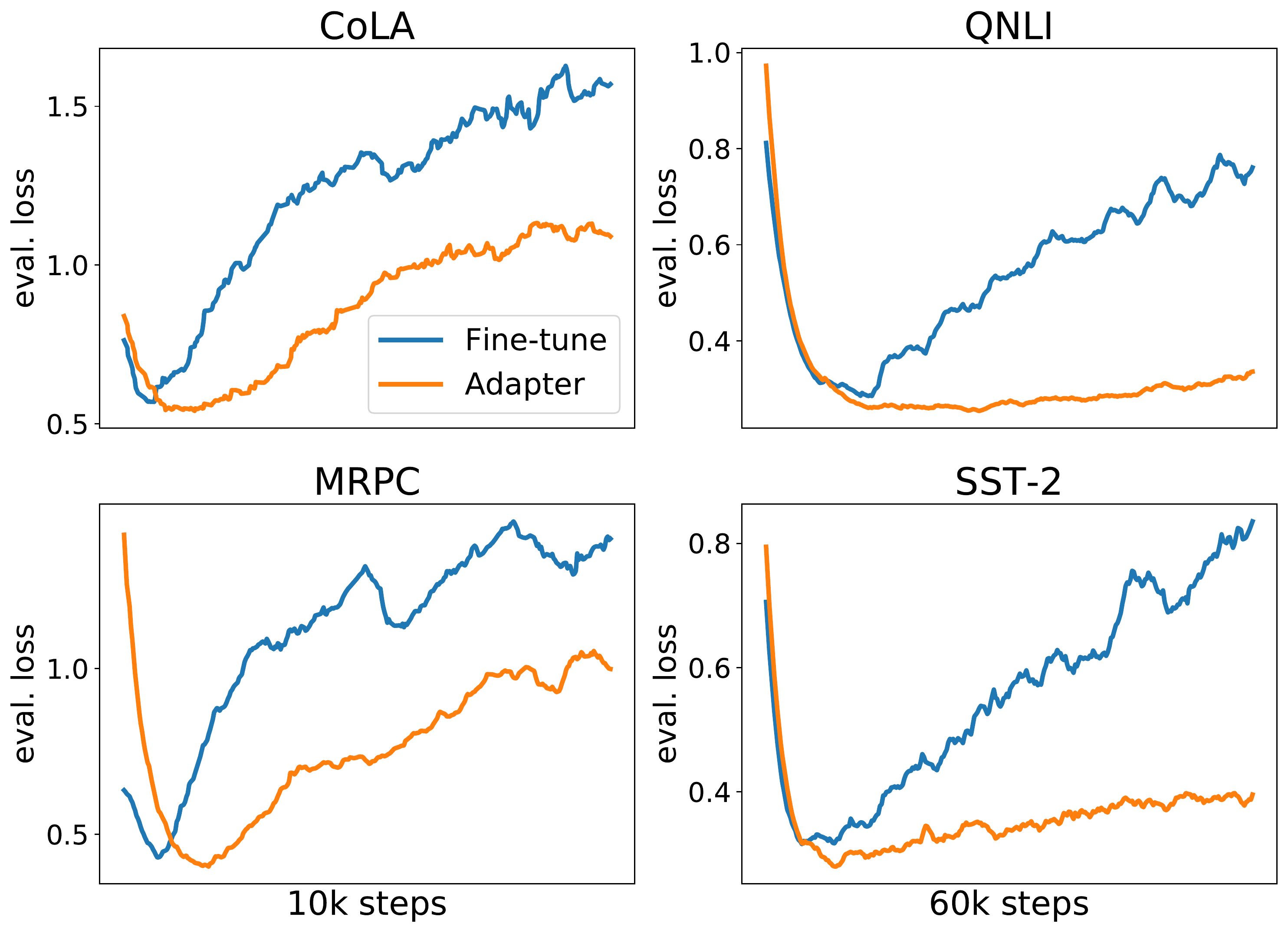}
\end{center}
\caption{Loss on the dev set w.r.t training steps. Results are based on BERT-base. The original training and dev sets from GLUE are used for this analysis. }
\label{fig:overfit_loss}
\end{figure}

\begin{table}[]
\centering
\scalebox{0.85}{
\begin{tabular}{lcc}
\toprule
\multirow{2}{*}{\textbf{Eval acc.}} & \multicolumn{2}{c}{\textbf{Mean~(Best)}} \\ \cline{2-3}
          & \textbf{Fine-tune}  & \textbf{Adapter}  \\ \hline
CoLA      & 54.27~(61.99)        & 58.27~(62.07)      \\ 
MRPC      & 84.53~(87.50)        & 85.28~(87.25)      \\ 
QNLI      & 89.39~(90.63)        & 90.41~(91.16)      \\ 
SST-2     & 90.21~(92.66)        & 91.01~(92.20)      \\ \bottomrule
\end{tabular}
}
\caption{Mean (Best) results on the dev set across all evaluation steps.}
\label{tab:overfit_result}
\end{table}

\paragraph{Overfitting and Generalization} 
Here, we first study the robustness of adapter-based tuning to overfitting. We use CoLA, MRPC, QNLI, and SST-2 with their original training and development sets for our analysis. The CoLA and MRPC contain 8.5k and 3.7k training samples and are regarded as low-resource tasks. The QNLI and SST-2 contain 104k and 67k training samples and are used as high-resource tasks. We train the two low-resource tasks for 10k steps, and the high resource tasks for 60k steps with a batch size of 16. We use BERT-base for all experiments. Figure~\ref{fig:overfit_loss} plots the loss curves on dev sets w.r.t training steps. We observe that models with fine-tuning can easily overfit on both low- and high-resource tasks. Adapter-based tuning is more robust to overfitting. Additional results on accuracy w.r.t. training steps and a similar analysis on XNLI are in Appendix~\ref{append: additional_results}. 

We also present the mean and best dev results across all evaluation steps in Table~\ref{tab:overfit_result}, where we perform an evaluation step every 20 training steps. The mean results of adapter-based tuning consistently outperform those of fine-tuning. The differences between the mean and the best values are also smaller with adapter-based tuning. The results suggest that the performance of adapters is more stable over fine-tuning along the training process.  

\begin{figure}[t]
\setlength{\abovecaptionskip}{-0.1cm}
\setlength{\belowcaptionskip}{0.1cm}
\begin{center}
\includegraphics[width=\columnwidth]{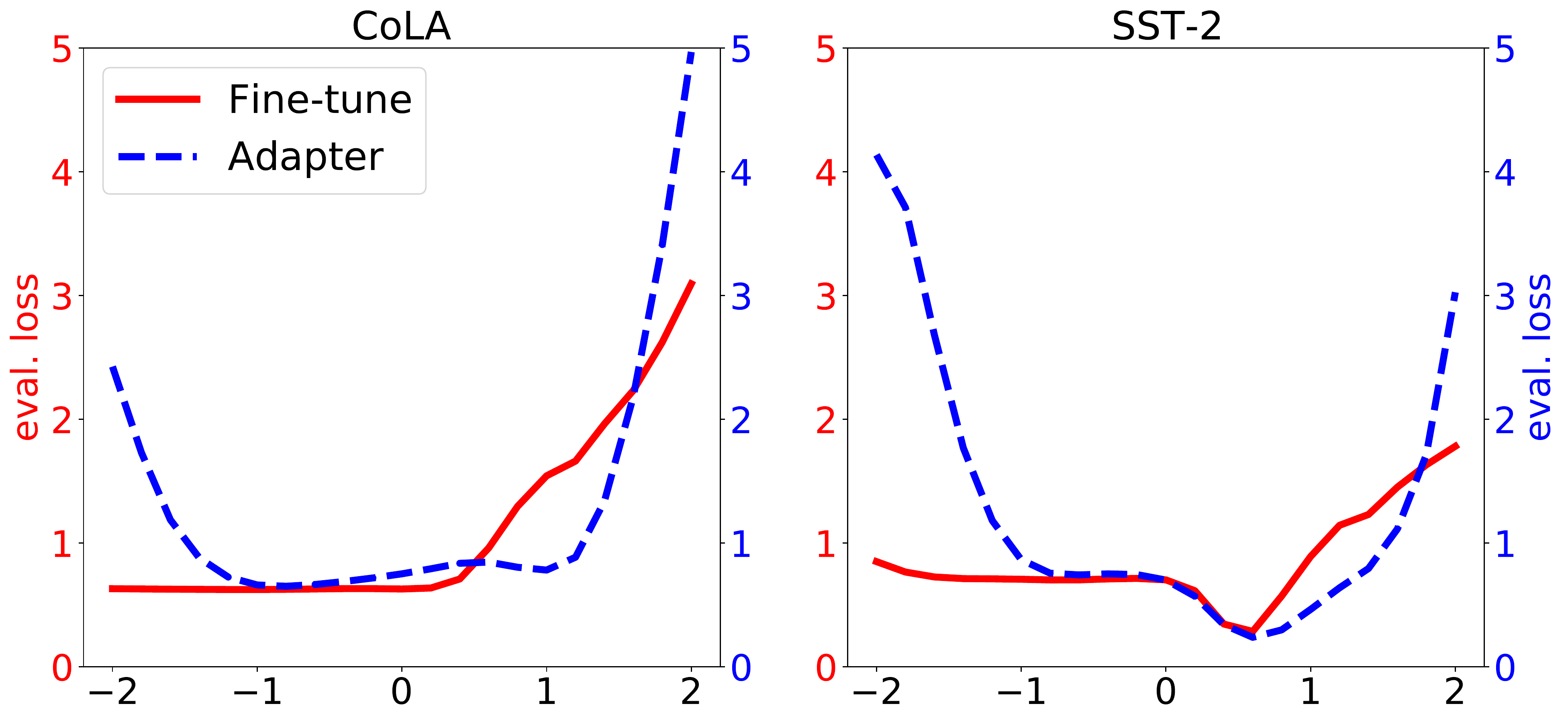}
\end{center}
\caption{Loss landscapes. BERT-base is used.}
\label{fig:loss_landscape}
\end{figure}

Training neural networks can be viewed as searching for a good minima in the non-convex landscape defined by the loss function. Prior work~\cite{Hochreiter1997flat,li2018visual} shows that the flatness of a local minima correlates with the generalization capability. 
Thus, we further show the loss landscapes of the two tuning methods. Following \citet{hao2019visualizing}, we plot the loss curve by linear interpolation between $\theta_0$ and $\theta_1$ with function $f(\alpha)=\mathcal{L}(\theta_0 + \alpha \cdot (\theta_1 - \theta_0))$, where $\theta_0$ and $\theta_1$ denote the model weights before and after tuning. $\mathcal{L}(\theta)$ is the loss function and $\alpha$ is a scalar parameter. In our experiments, we set the range of $\alpha$ to $[-2,2]$ and uniformly sample 20 points. Figure~\ref{fig:loss_landscape} shows the loss landscape curves on CoLA and SST based on BERT-base. It shows that the minimas of adapter-based tuning are more wide and flat, which indicates that adapter-based tuning tends to generalize better.

\paragraph{Compare to Mixout} 

\begin{table}[]
\centering
\scalebox{0.82}{
\begin{tabular}{lcccc}
\toprule
\textbf{Model} &\bf{CoLA} &\bf{MRPC} &\bf{QNLI} &\bf{SST-2}\\
\midrule
finetune &41.39 &83.56 &80.51 &87.84\\
finetune-mixout &42.35 &84.00 &80.03 &87.71 \\

Adapter$_{64}$ &\bf{42.93} &\bf{84.79} &80.54 &\bf{88.02}\\
Adapter$_{64}$-mixout &42.52 &83.80 &\bf{80.67} &87.66 \\
\bottomrule
\end{tabular}}
\caption{Comparison with Mixout. Results are based on BERT-base under 1k settiing. Average results across 5 random seeds are reported.} \label{table:mixout}
\end{table}
The focus of this paper is to answer the question -- besides being parameter-efficient, when would adapter-based tuning be more effective than fine-tuning for PrLM adaptation? Thus, we only use fine-tuning as our primary baseline in previous sections. Here, for the sake of curiosity, we further compare adapter-based tuning to fine-tuning regularized by \emph{mixout}~\cite{{DBLP:conf/iclr/LeeCK20}} on a subset of GLUE tasks, since \emph{mixout} similarly regularizes the learning process by mitigating the forgetting issue. Specifically, it replaces all outgoing parameters from a randomly selected neuron to the corresponding parameters of the initial model without tuning, such that it reduces divergence from the initial model. Following the suggestions in the paper, we conduct experiments by replacing all \emph{dropout} modules in the network with \emph{mixout} and set the mixout probability to $0.9$. 
From the results in Table~\ref{table:mixout}, we find that using adapter-based tuning alone yields the best results in most cases. Applying \emph{mixout} to fine-tuning improves the performance on CoLA and MRPC only. However, applying it to adapters instead tends to degrade the performance. We suspect that this is because the number of trainable parameters of adapters is very few to begin with. Hence, further replacing a large percentage of them with their initial weights may weaken the learning ability. 


\section{Related Work}

Fine-tuning pretrained large scale language models has proven its effectiveness on a wide range of NLP tasks~\citep{devlin2019bert,liu2019roberta,conneau2020unsupervised,brown2020language}. However, fine-tuning requires a new set of weights for each task, which is parameter inefficient. Adapter-based tuning is proposed to deal with this problem~\cite{houlsby2019a}. Most previous work has demonstrated that it achieves comparable performance to fine-tuning~\cite{bapna2019simple,DBLP:conf/emnlp/PfeifferRPKVRCG20,pfeiffer2020adapterfusion,pfeiffer2020madx,ruckle2020adapterdrop,wang2020kadapter,DBLP:conf/nips/GuoZXWCC20}. However, existing work mostly focuses on the parameter-efficient aspect while overlooks the effectiveness. 

Fine-tuning PrLMs in a low-resource setting has been studied for a while~\cite{DBLP:journals/corr/abs-2002-06305,DBLP:conf/iclr/LeeCK20,DBLP:journals/corr/abs-1811-01088,DBLP:conf/acl/JiangHCLGZ20,DBLP:journals/corr/abs-2006-05987}. Previous work points out that with large-scale parameters, fine-tuning on a few samples can lead to overfitting and bad generalization, which causes the results unstable. \citet{DBLP:journals/corr/abs-1811-01088} find that pretraining on an intermediate task can improve fine-tuning outcomes.  \citet{DBLP:conf/acl/JiangHCLGZ20} improve the robustness of fine-tuning by controlling the model complexity and preventing aggressive updating.
On the other hand, catastrophic forgetting can appear when transferring a pretrained neural networks~\cite{french1999catastrophic,mccloskey1989catastrophic,goodfellow2013empirical}, where the learned knowledge from pretraining is lost when adapting to downstream tasks. This phenomenon often appears in NLP tasks~\cite{DBLP:conf/emnlp/MouMYLX0J16,DBLP:conf/acl-alta/AroraRB19}. To relieve this problem of adapting pretrained language models, \citet{DBLP:conf/acl/RuderH18} gradually unfreeze the layers starting from the last layer and \citet{DBLP:conf/cncl/SunQXH19} find assigning lower learning rate to the bottom layers can improve the performance. \citet{DBLP:conf/iclr/LeeCK20} regularize learning by encouraging the weights of the updated model to stay close to the initial weights. \citet{aghajanyan2020better} regularize fine-tuning by introducing noise to the input which is similar to adversarial training for fine-tuning studied in \citet{DBLP:conf/iclr/ZhuCGSGL20}. \citet{DBLP:journals/corr/abs-2006-04884} point out that the instability of fine-tuning lies in the optimizer and propose to revise the Adam optimizer by replacing it with a de-bias version. \citet{DBLP:conf/emnlp/ChenHCCLY20} propose a mechanism to recall the knowledge from pretraining tasks.

\section{Conclusion}\label{sec:conclusion}
Prior work often focuses on the parameter-efficient aspect while overlooks the effectiveness of adapter-based tuning. We empirically demonstrate that adapter-based tuning can better regularize the learning process. We conduct extensive experiments to verify its effectiveness and conclude that 1) it tends to outperform fine-tuning on both low-resource and cross-lingual tasks; 2) it demonstrates higher stability under different learning rates compared to fine-tuning. We hope our study will inspire more future work on PrLM adaptation based on adapters and other methods that only tune part of the PrLM parameters.

\section*{Acknowledgements}
\label{sec:acknowledgements}

Linlin Liu would like to thank the support from Interdisciplinary Graduate School, Nanyang Technological University.

\bibliographystyle{acl_natbib}
\bibliography{anthology,acl2021}

\clearpage
\pagenumbering{arabic}

\appendix

\section{Appendix}
\subsection{Datasets} \label{append:data}

\paragraph{TAE} 
Table~\ref{table: tapt_data_statistics} presents the data statistics of the TAE datasets we used in $\S$~\ref{sec:tapt}.

\paragraph{GLUE} 
Table~\ref{table: glue_data_statistics} presents the statistics and descriptions of GLUE tasks. In $\S$~\ref{sec:glue}, to investigate the effectiveness in low-resource scenarios,  we simulate two low-resource settings by randomly sampling 1k and 5k examples respectively from each of the original training set as the new training sets.  In each setting, we draw 1k samples from the remaining training set as our validation set and use the original validation set as held-out test set since the original GLUE test sets are not publicly available.

For the RSA analysis in $\S$~\ref{sec:model} and the analysis of overfitting and generalization in $\S$~\ref{sec:analysis}, we use the original training and development sets for analysis purpose, as this better reveals the behaviors under both high- and low- resource settings. 
\bigskip

\subsection{Experimental Details}\label{append: experimental setup}
\paragraph{Implementation}
We use language model implementations from \emph{HuggingFace Transfromers} library~\cite{Wolf2019}. Our adapter implementation is also based on that. Following standard practice~\cite{devlin2019bert}, we pass the final layer [CLS] token representation to a task-specific feedforward layer for prediction on downstream tasks. Each experiment was performed on a single v100 GPU. We use the Adam optimizer~\cite{Kingma2015} with a linear learning rate scheduler. 

\paragraph{Training Details on TAE and GLUE}
For both fine-tuning and adapter-based tuning, we train models for a fixed number of epochs, and select models with the best validation performances on epoch end for evaluation. 

For fine-tuning, on TAE we follow the learning rate and batch size as suggested by~\citet{houlsby2019a}. On GLUE, we tune learning rates in \{1e-5, 2e-5, 3e-5, 4e-5, 5e-5\} and batch sizes in \{16, 32\} to select the best configuration across tasks. 

For adapters, on TAE, we set the batch size the same as used in fine-tuning, and tune learning rates in \{2e-5, 5e-5,1e-4, 2e-4\} and adapter's hidden size in \{64, 128, 256\} to select the best configuration across all tasks. On GLUE, we keep the learning rate and batch size the same as used in TAE, and tune the adapter's hidden sizes in \{64, 128, 256\} for each task.  We use the same hyperparameter setting for all our analysis experiments with GLUE tasks as well. 

Table~\ref{table: text_clf_hyperparam} presents the detailed hyperparameter settings for TAE and GLUE.

\paragraph{Training Details on Xtreme Tasks}
For UD-POS, Wikiann NER, and XNLI, we use batch size 32, and tune learning rates in \{1e-5, 2e-5, 3e-5,  4e-5,  5e-5\} on each task. We tune the adapter's hidden sizes in \{64, 128, 256\} to select the best value across all tasks. We use the English training and development sets of each task for hyperparameter tuning. Table~\ref{table: xtreme_clf_hyperparam} presents the detailed settings.
\bigskip

\subsection{Additional Results}~\label{append: additional_results}

\begin{figure*}[]
\setlength{\abovecaptionskip}{-0.1cm}
\setlength{\belowcaptionskip}{-0.5cm}
\begin{center}
\includegraphics[width=0.9\textwidth]{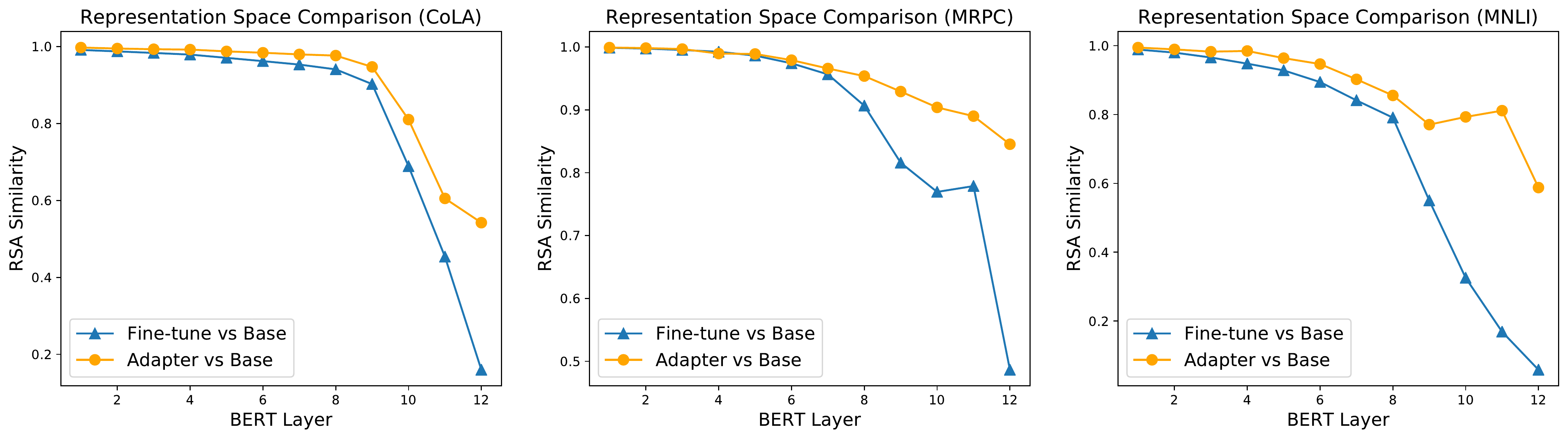}
\end{center}
\caption{Comparison of the representations obtained at each layer before (\emph{Base}) and after adapter-based tuning or fine-tuning on BERT-base using Representational Similarity Analysis (RSA). The original training and dev sets of CoLA, MRPC, and MNLI are used for this analysis. 5000 tokens are randomly sampled from the dev set of each task for computing RSA. A higher score indicates that the representation spaces before and after tuning are more similar.}
\label{fig:rsa_glue}
\end{figure*}

\begin{figure*}[]
\setlength{\abovecaptionskip}{-0.1cm}
\setlength{\belowcaptionskip}{-0.5cm}
\begin{center}
\includegraphics[width=0.9\textwidth]{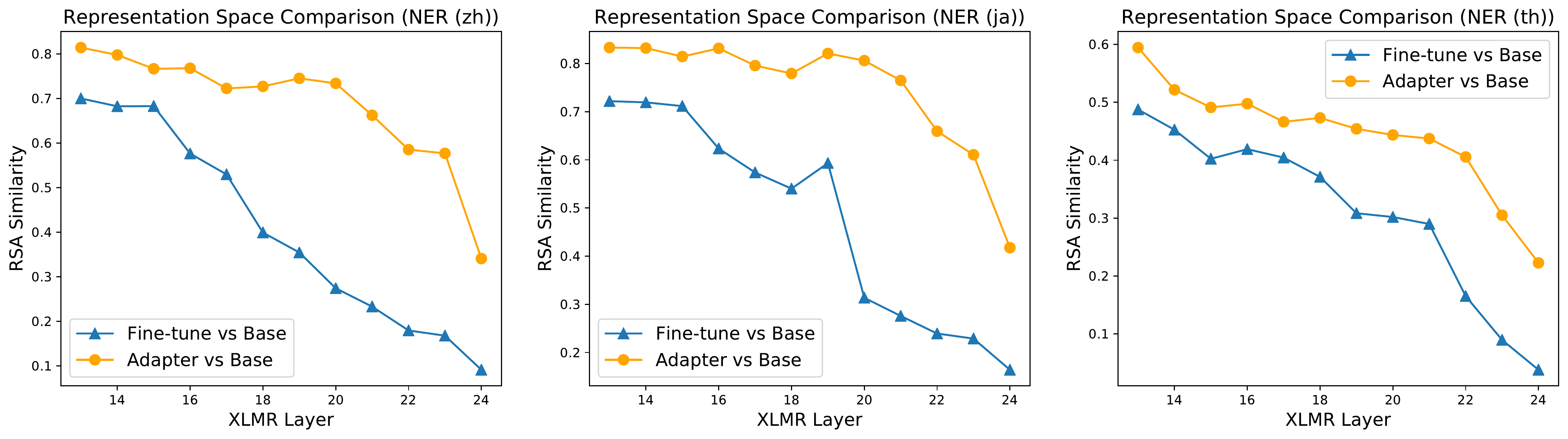}
\end{center}
\caption{Comparison of the representations obtained at each of the top 12 layers (layer 13-24) before (\emph{Base}) and after adapter-based tuning or fine-tuning on XLMR-large using Representational Similarity Analysis (RSA). We show results on 3 distant languages from the Wikiann NER task. 5000 tokens are randomly sampled from the dev set of each language for computing RSA. A higher score indicates that the representation spaces before and after tuning are more similar.}
\label{fig:rsa_xtreme}
\end{figure*}

\paragraph{RSA}
Figure~\ref{fig:rsa_glue} presents additional Representational Similarity Analysis (RSA) plots on three GLUE tasks as mentioned in $\S$~\ref{sec:model}. We further conduct RSA to show the deviation of representation space before and after tuning (with English training set) on three distant languages (zh, ja, th) from the cross-lingual NER task. Figure~\ref{fig:rsa_xtreme} presents the results.

\begin{figure*}[]
\setlength{\abovecaptionskip}{-0.1cm}
\setlength{\belowcaptionskip}{-0.5cm}
\begin{center}
\includegraphics[width=\textwidth]{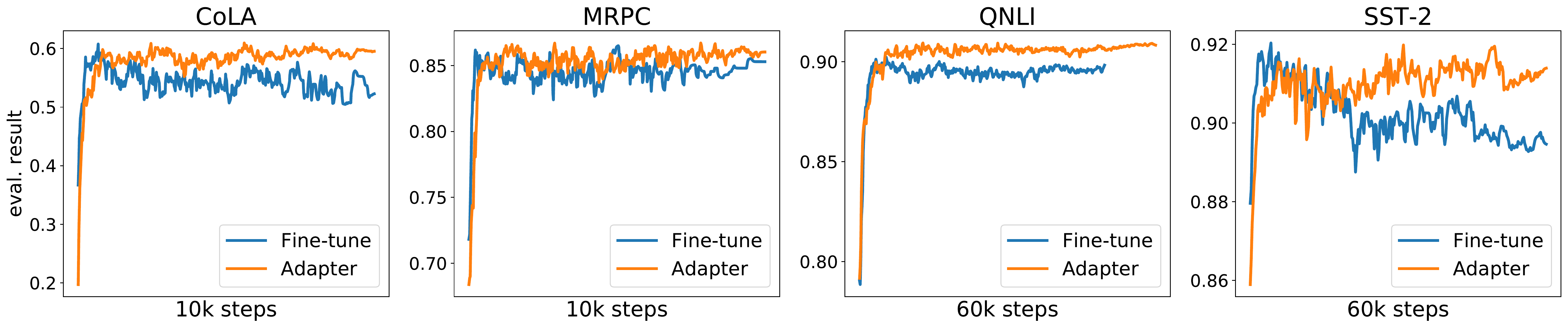}
\end{center}
\caption{Accuracy on the dev set w.r.t training steps. Results are based on BERT-base. The original training and dev sets from GLUE are used for this analysis. We can observe that for both high resource (QNLI and SST-2) and low-resource (CoLA and MRPC) tasks, adapter-based tuning is more robust to overfitting.}
\label{fig:overfit_acc}
\end{figure*}

\paragraph{Accuracy w.r.t Training Steps} 
Figure~\ref{fig:overfit_acc} shows the change of accuracy with increasing training steps on four GLUE tasks. The results again indicate that adapter-based tuning is more robust to overfitting.

\begin{figure*}[]
\setlength{\abovecaptionskip}{-0.1cm}
\setlength{\belowcaptionskip}{-0.5cm}
\begin{center}
\includegraphics[width=0.7\textwidth]{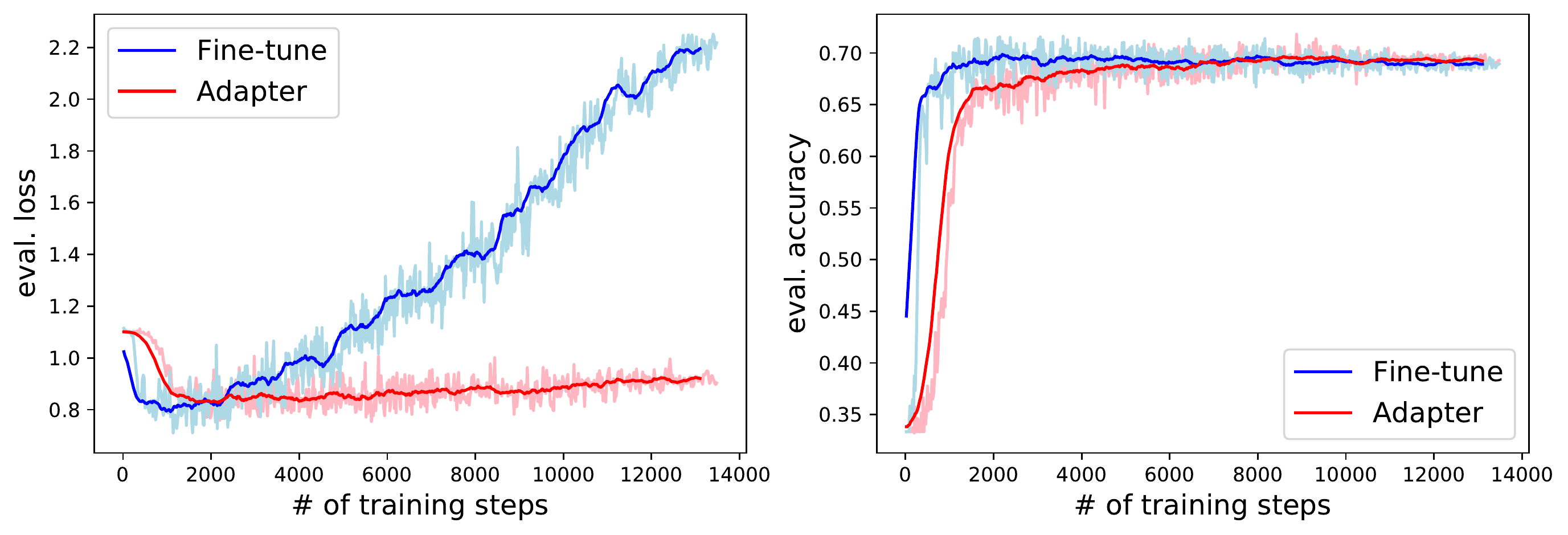}
\end{center}
\caption{Change of the average dev loss (left) and accuracy (right) across all target languages of XNLI except English with increasing training steps. The results are obtained when trained on 10\% of the XNLI training data.}
\label{fig:overfit_xtreme}
\end{figure*}

\paragraph{Overfitting Analysis on XNLI}
We train XLMR-large with 10\% of the original English training data of XNLI, and plot the average loss and accuracy curves on development sets across all target languages except English in Figure~\ref{fig:overfit_xtreme}. The plots demonstrate similar trends as shown in the plots of GLUE tasks (Figure~\ref{fig:overfit_loss} and Figure~\ref{fig:overfit_acc}), where models with fine-tuning are easily overfitted and adapter-based tuning is more robust to overfitting.

\paragraph{Detailed Cross-lingual Results}
Table~\ref{table:pos_appendix} and Table~\ref{table:ner_appendix} presents the cross-lingual POS tagging results and the cross-lingual NER results on each language respectively. Table~\ref{table:xnli_high_resource_details} presents detailed results on XNLI when trained with full data.
\ref{table:xnli_low_resource_details} presents detailed XNLI results when trained on 5\%, 10\%, and 20\% of training data. 

\begin{table*}[t]
\centering
\scalebox{0.9}{
\begin{tabular}{lllrrrr}
\toprule
\bf{Domain} &\bf{Task} &\bf{Label Type} &\bf{\# Train} &\bf{\# Dev} &\bf{\# Test} &\bf{\# Class} \\ \midrule

\multirow{2}*{BIOMED} &CHEMPROT &relation classification &4169 &2427 &3469 &13\\
&RCT &abstract sent. roles &180040 &30212 &30135 &5\\
\midrule

\multirow{2}*{CS} &ACL-ARC &citation intent &1688  &114 &139 &6\\
&SCIERC &relation classification &3219 &455 &974 &7\\
\midrule

\multirow{2}*{NEWS} &HYPERPARTISAN &partisanship &515 &65 &65 &2\\
&AGNEWS &topic &115000 &5000 &7600 &4\\
\midrule

\multirow{2}*{REVIEWS} &HELPFULNESS &review helpfulness &115251 &5000 &25000 &2 \\
&IMDB &review sentiment &20000 &5000 &25000 &2 \\

\bottomrule
\end{tabular}}
\caption{Data statistics of Task Adaptation Evaluation (TAE) tasks.} \label{table: tapt_data_statistics}
\end{table*}
\begin{table*}[t]
\centering
\scalebox{0.95}{
\begin{tabular}{llrcc}
\toprule
\bf{Task} &\bf{Description} &\bf{\# Train} &\bf{\# Dev} &\bf{\# Class} \\ \midrule

CoLA &linguistic acceptability classification&8.5k &1042 &2 \\
MNLI &textual entailment classification&392k &9816/9833 &3 \\
MRPC &paraphrase classification &3.7k &409 &2 \\
QNLI &textual entalment classification&104k &5464 &2 \\
QQP &quora question paris classification &363k &404k &2 \\
RTE &textual entailment classification&2.5k &278 &2 \\
SST-2 &sentiment classification &67k &873 &2 \\
STS-B &semnatic textual similarity (regression) &5.7k &1501 &- \\
\bottomrule
\end{tabular}}
\caption{Data statistics of GLUE tasks.} \label{table: glue_data_statistics}
\end{table*}
\begin{table*}[t]
\centering
\scalebox{0.9}{
\begin{tabular}{lcc @{\hspace{1.5cm}} cc}
\toprule
&\multicolumn{2}{c}{\bf{TAE}} &\multicolumn{2}{c}{\bf{GLUE}}\\
\bf{Hyperparameter}&\bf{fine-tuning} &\bf{adapter} &\bf{fine-tuning} &\bf{adapter} \\
\midrule
number of epochs &10 &10 &20 &20\\
batch size &16 &16 &16 &16\\
learning rate &2e-5 &1e-4 &2e-5 &1e-4\\
dropout &0.1 &0.1 &0.1 &0.1\\
feedforward layer &1 &1 &1 &1\\
feedforward nonlnearity layer &1 &1 &1 &1\\
classification layer &1 &1 &1 &1\\

\bottomrule
\end{tabular}}
\caption{Hyperparameters for fine-tuning and adapter-based tuning for experiments on TAE and GLUE.} \label{table: text_clf_hyperparam}
\end{table*}
\begin{table*}[t]
\centering
\scalebox{0.8}{
\begin{tabular}{lcc @{\hspace{1.2cm}} cc @{\hspace{1.2cm}} cc}
\toprule
&\multicolumn{2}{c}{\bf{POS}} &\multicolumn{2}{c}{\bf{NER}} &\multicolumn{2}{c}{\bf{XNLI}}\\

\bf{Hyperparameter} &\bf{fine-tune} &\bf{adapter} &\bf{fine-tune} &\bf{adapter} &\bf{fine-tune} &\bf{adapter} \\
\midrule
number of epochs &5 &5 &5 &5 &5 &5\\
batch size &32 &32 &32 &32 &32 &32\\
learning rate &2e-5 &5e-5 &2e-5 &5e-5 &1e-5 &4e-5\\
dropout &0.1 &0.1 &0.1 &0.1 &0.1 &0.1\\
feedforward layer &1 &1 &1 &1 &1 &1\\
feedforward nonlnearity layer &1 &1 &1 &1 &1 &1\\
classification layer &1 &1 &1 &1 &1 &1\\

\bottomrule
\end{tabular}}
\caption{Hyperparameters for fine-tuning and adapter-based tuning for experiments on UD-POS, Wikiann NER, and XNLI.} \label{table: xtreme_clf_hyperparam}
\end{table*}

\begin{table*}[t]
\centering
\scalebox{0.6}{
\begin{tabular}{lccccccccccccccccc}
\toprule
 & \textbf{en} & \textbf{af} & \textbf{ar} & \textbf{bg} & \textbf{de} & \textbf{el} & \textbf{es} & \textbf{et} & \textbf{eu} & \textbf{fa} & \textbf{fi} & \textbf{fr} & \textbf{he} & \textbf{hi} & \textbf{hu} & \textbf{id} & \textbf{it} \\
\midrule
Indo-European & yes & yes & no & yes & yes & yes & yes & no & no & yes & no & yes & no & yes & no & no & yes \\
XLMR-ft$^\dag$ & 96.10 &	89.80 &	67.50 &	88.10 &	88.50 &	86.30 &	88.30 &	86.50 &	72.50 &	70.60 &	85.80 &	87.20 &	68.30 &	56.80 &	82.60 &	72.40 &	89.40 \\
XLMR-ft$^*$ & 96.15 &	89.26 &	69.12 &	88.33 &	88.79 &	87.42 &	88.34 &	87.38 &	73.70 &	71.05 &	86.56 &	87.24 &	67.86 &	75.48 &	83.49 &	72.67 &	89.07 \\
XLMR-adapter$_{256}$ & 95.89 &	89.30 &	70.50 &	88.79 &	88.48 &	86.44 &	88.99 &	87.31 &	74.84 &	71.94 &	85.99 &	88.74 &	67.32 &	69.63 &	83.11 &	73.31 &	90.16 \\
\midrule
\midrule
 & \textbf{ja} & \textbf{kk} & \textbf{ko} & \textbf{mr} & \textbf{nl} & \textbf{pt} & \textbf{ru} & \textbf{ta} & \textbf{te} & \textbf{th} & \textbf{tl} & \textbf{tr} & \textbf{ur} & \textbf{vi} & \textbf{yo} & \textbf{zh} & \textbf{avg} \\
 \midrule
Indo-European & no & no & no & yes & yes & yes & yes & no & no & no & no & no & yes & no & no & no & -\\
XLMR-ft$^\dag$ & 15.90 & 78.10 & 53.90 &	80.80 &	89.50 &	87.60 &	89.50 &	65.20 &	86.60 &	47.20 &	92.20 &	76.30 &	70.30 &	56.80 &	24.60 &	25.70 & 73.80 \\
XLMR-ft$^*$ & 21.34 &	78.86 &	53.84 &	85.24 &	89.75 &	87.98 &	89.75 &	64.34 &	85.65 &	43.12 &	93.03 &	76.65 &	69.43 &	58.10 &	23.92 &	28.60 & 74.29 \\
XLMR-adapter$_{256}$ & 38.53 &	78.47 &	53.35 &	86.45 &	89.86 &	88.82 &	90.21 &	64.31 &	85.38 &	55.88 &	91.10 &	76.21 &	63.46 &	59.38 &	24.28 &	55.76 & 75.82 \\
\bottomrule
\end{tabular}}
\caption{Zero-shot cross-lingual POS tagging accuracy on the test set of each target language. Results with ``$\dag$'' are taken from~\citep{hu2020b}. Results with ``$*$'' are reproduced by us.}
\label{table:pos_appendix}
\end{table*}
\begin{table*}[t]
\centering
\scalebox{0.55}{
\begin{tabular}{lccccccccccccccccccccc}
\toprule
 & \textbf{en} & \textbf{ar} & \textbf{he} & \textbf{vi} & \textbf{id} & \textbf{jv} & \textbf{ms} & \textbf{tl} & \textbf{eu} & \textbf{ml} & \textbf{ta} & \textbf{te} & \textbf{af} & \textbf{nl} & \textbf{de} & \textbf{el} & \textbf{bn} & \textbf{hi} & \textbf{mr} & \textbf{ur} \\
\midrule
Indo-European & yes & no & no & no & no & no & no & no & no & no & no & no & yes & yes & yes & yes & yes & yes & yes & no \\
XLMR-ft$^\dag$ & 84.7 & 53 & 56.8 & 79.4 & 53 & 62.5 & 57.1 & 73.2 & 60.9 & 67.8 & 59.5 & 55.8 & 78.9 & 84 & 78.8 & 79.5 & 78.8 & 73 & 68.1 & 56.4 \\
XLMR-ft$^*$ & 84.62 & 43.72 & 54.08 & 77.19 & 52.26 & 58.37 & 69.78 & 72.21 & 62.08 & 65.78 & 56.92 & 52.31 & 77.64 & 84.26 & 77.95 & 77.23 & 76.25 & 71.01 & 64.14 & 54.15 \\
XLMR-adapter$_{256}$ & 83.87 & 51.89 & 56.59 & 78.02 & 53.53 & 63.24 & 62.65 & 71.57 & 64.96 & 68.30 & 59.57 & 54.93 & 79.43 & 84.88 & 79.38 & 80.51 & 78.99 & 73.17 & 72.74 & 72.36  \\
\midrule
\midrule
 & \textbf{fa} & \textbf{fr} & \textbf{it} & \textbf{pt} & \textbf{es} & \textbf{bg} & \textbf{ru} & \textbf{ja} & \textbf{ka} & \textbf{ko} & \textbf{th} & \textbf{sw} & \textbf{yo} & \textbf{my} & \textbf{zh} & \textbf{kk} & \textbf{tr} & \textbf{et} & \textbf{fi} & \textbf{hu}  \\
 \midrule
Indo-European & yes & yes & yes & yes & yes & yes & yes & no & no & no & no & no & no & no & no & no & no & no & no & no \\
XLMR-ft$\dag$ & 61.9 & 80.5 & 81.3 & 81.9 & 79.6 & 81.4 & 69.1 & 23.2 & 71.6 & 60 & 1.3 & 70.5 & 33.6 & 54.3 & 33.1 & 56.2 & 76.1 & 79.1 & 79.2 & 79.8 \\
XLMR-ft$^*$ & 61.13 & 79.07 & 81.05 & 79.61 & 68.76 & 81.18 & 71.46 & 18.31 & 68.93 & 57.99 & 1.47 & 69.95 & 41.26 & 51.32 & 25.82 & 49.83 & 78.94 & 78.03 & 78.63 & 79.32 \\
XLMR-adapter$_{256}$ & 60.39 & 81.21 & 81.79 & 82.61 & 76.12 & 82.50 & 69.76 & 21.41 & 70.55 & 61.37 & 2.47 & 68.90 & 38.18 & 60.48 & 31.11 & 51.34 & 81.89 & 80.36 & 80.86 & 82.06 \\
\bottomrule
\end{tabular}}
\caption{Zero-shot cross-lingual NER F1 on the test set of each language. Results with ``$\dag$'' are taken from~\citep{hu2020b}. Results with ``$*$'' are reproduced by us.}
\label{table:ner_appendix}
\end{table*}
\begin{table*}[ht]
\centering
\scalebox{0.7}{
\begin{tabular}{lccccccccccccccc}
\toprule
Model & \textbf{en} & \textbf{ar} & \textbf{bg} & \textbf{de} & \textbf{el} & \textbf{es} & \textbf{fr} & \textbf{hi} & \textbf{ru} & \textbf{sw} & \textbf{th} & \textbf{tr} & \textbf{ur} & \textbf{vi} & \textbf{zh} \\
\midrule
XLMR-ft$^\dag$ & 88.7 & 77.2 & 83 & 82.5 & 80.8 & 83.7 & 82.2 & 75.6 & 79.1 & 71.2 & 77.4 & 78.0 & 71.7 & 79.3 & 78.2 \\
XLMR-ft$^*$ &88.28 &78.34 &82.73 &82.07 &81.34 &83.63 &81.93 &75.33 &79.04 &71.59 &76.67 &78.36 &71.86 &79.32 &78.80\\
XLMR-adapter$_{256}$ & 89.22 & 78.62 & 83.59 & 83.47 & 82.39 & 84.69 & 83.27 & 76.42 & 79.74 & 72.21 & 77.84 & 78.80 & 72.27 & 79.32 & 79.34 \\
\bottomrule
\end{tabular}}
\caption{Zero-shot XNLI accuracy on the test set of each language when trained with full data. Results with ``$\dag$'' are taken from~\citep{hu2020b}. Results with ``$*$'' are reproduced by us.}
\label{table:xnli_high_resource_details}
\end{table*}

\begin{table*}[ht]
\centering
\scalebox{0.7}{
\begin{tabular}{lccccccccccccccc}
\toprule
Model & \textbf{en} & \textbf{ar} & \textbf{bg} & \textbf{de} & \textbf{el} & \textbf{es} & \textbf{fr} & \textbf{hi} & \textbf{ru} & \textbf{sw} & \textbf{th} & \textbf{tr} & \textbf{ur} & \textbf{vi} & \textbf{zh} \\
\midrule
\textbf{5\% training data} \\
XLMR-ft & 85.09 & 73.53 & 78.7 & 79.58 & 77.26 & 80.13 & 79.36 & 72.07 & 76.52 & 67.8 & 72.53 & 74.53 & 68.3 & 75.24 & 75.74 \\
XLMR-adapter$_{64}$ & 84.77 & 73.95 & 78.76 & 79.02 & 78.08 & 80.55 & 79.48 & 72.01 & 76.54 & 68.76 & 73.83 & 75.56 & 68.6 & 75.94 & 75.50 \\
\midrule
\textbf{10\% training data} \\
XLMR-ft & 85.96 & 75.04 & 79.78 & 79.82 & 78.72 & 80.99 & 80.25 & 73.23 & 77.28 & 68.08 & 74.43 & 75.7 & 69.54 & 76.02 & 76.16 \\
XLMR-adapter$_{64}$ & 85.74 & 76.78 & 80.27 & 80.77 & 79.72 & 81.87 & 81.13 & 73.87 & 78.42 & 69.3 & 74.25 & 77.08 & 69.54 & 77.30 & 76.82 \\
\midrule
\textbf{20\% training data} \\
XLMR-ft & 87.26 & 76.48 & 81.07 & 82.03 & 80.47 & 82.55 & 81.53 & 75.06 & 78.04 & 69.96 & 76.00 & 77.36 & 70.75 & 77.74 & 77.94 \\
XLMR-adapter$_{64}$ & 87.24 & 78.00 & 81.87 & 82.15 & 80.47 & 82.65 & 81.53 & 75.00 & 78.74 & 70.87 & 75.94 & 78.44 & 70.51 & 78.70 & 78.16 \\
\bottomrule
\end{tabular}}
\caption{Zero-shot XNLI accuracy on the test set of each language when trained on 5\%, 10\%, 20\% of training data respectively.}
\label{table:xnli_low_resource_details}
\end{table*}

\end{document}